\documentclass[letterpaper]{article} 
\usepackage{aaai25}  
\usepackage{cite}
\usepackage{amsmath,amssymb,amsfonts}
\usepackage{algorithmic}
\usepackage{graphicx}
\usepackage{textcomp}
\usepackage{xcolor}
\usepackage{booktabs}
\usepackage{multirow}
\usepackage{array}
\usepackage{hyperref}
\usepackage{subcaption}

\title{FastAT Benchmark: A Comprehensive Framework for Fair Evaluation of Fast Adversarial Training Methods}

\author {
    Chao Pan\textsuperscript{\rm 1,2},
    Xin Yao\textsuperscript{\rm 3}
}
\affiliations {
    \textsuperscript{\rm 1}Department of Computer Science and Engineering,\\
    Southern University of Science and Technology, Shenzhen 518055, China\\
    \textsuperscript{\rm 2}The Hong Kong Polytechnic University, Hong Kong, China\\
    \textsuperscript{\rm 3}School of Data Science, Lingnan University, Hong Kong, China\\
    11930665@mail.sustech.edu.cn
}

\begin{document}
\maketitle

\begin{abstract}
Fast Adversarial Training (FastAT) seeks to achieve adversarial robustness at a fraction of the computational cost incurred by standard multi-step methods such as PGD-AT. Although numerous FastAT techniques have been proposed in recent years, fair comparison among them remains elusive. Existing benchmarks and public leaderboards typically permit diverse model architectures, varying training configurations, and external data sources, making it unclear whether reported improvements reflect genuine algorithmic advances or merely more favorable experimental conditions. To address this problem, we introduce the FastAT Benchmark, a controlled evaluation framework built on three core design principles: unified architecture requirements, standardized training settings, and strict prohibition of external or synthetic data. The benchmark implements over twenty representative FastAT methods within a single codebase, enabling direct and reproducible comparison. Each method is assessed through a dual-metric evaluation framework that measures both adversarial robustness (accuracy under PGD, AutoAttack, and CR Attack) and computational cost (GPU training time and peak memory footprint). Comprehensive experiments on CIFAR-10, CIFAR-100, and Tiny-ImageNet provide reliable baseline measurements and reveal that well-designed single-step methods can match or surpass PGD-AT robustness at substantially lower cost, while no single method dominates across all evaluation dimensions. The complete benchmark, including source code, configuration files, and experimental results, is publicly available to support transparent and fair evaluation of future FastAT research.\footnote{Source code is available at \url{https://github.com/fzjcdt/FastAT_Benchmark}. Project homepage: \url{https://fzjcdt.github.io/FastAT_Benchmark/benchmark.html}.}

\end{abstract}


\section{Introduction}
\label{sec:introduction}
Adversarial training has established itself as one of the most effective defenses against adversarial examples~\cite{pgd}. Standard multi-step methods such as PGD-AT~\cite{pgd}, however, generate adversarial perturbations through multiple forward and backward passes at each training iteration, incurring substantial computational overhead. This overhead becomes prohibitive for large-scale applications, motivating extensive research into Fast Adversarial Training (FastAT) methods that seek to achieve comparable robustness at significantly reduced training cost.

The FastAT field has progressed rapidly, producing a diverse array of algorithmic innovations. These methods span several broad categories: randomized perturbation initialization approaches such as FGSM-RS~\cite{fgsm_rs}, N-FGSM~\cite{n_fgsm}, and ZeroGrad~\cite{zero_grad}; informed initialization techniques including FGSM-PGI~\cite{fgsm_pgi}, FGSM-UAP~\cite{fgsm_uap}, and SSAT~\cite{ssat}; gradient regularization methods such as GradAlign~\cite{grad_align} and GAT~\cite{gat}; output consistency regularization approaches including NuAT~\cite{nu_at} and AAER~\cite{aaer}; and various hybrid strategies~\cite{fgsm_mep_cs_fgsm_rs_cs, fgsm_pco, elle, liet}. While this diversity reflects the vitality of the field, it also raises a fundamental question: \textit{how can these methods be compared in a fair and meaningful way?}

Despite the growing number of proposed methods, the evaluation of FastAT techniques faces what we term a \emph{crisis of comparability}. Public leaderboards such as RobustBench~\cite{robustbench} have played a valuable role in establishing common evaluation standards for adversarial robustness, yet current leaderboard practices suffer from several limitations. First, top-performing entries employ vastly different model architectures, ranging from ResNets~\cite{resnet} to Vision Transformers~\cite{real_vit} and WideResNets~\cite{wide_resnet}. These architectural differences make it difficult to determine whether a reported improvement originates from a better training algorithm or simply from a more expressive model. Second, substantial variations in training duration, optimizers, learning rate schedules, and data augmentation strategies further obscure whether performance gains reflect genuine algorithmic innovation or merely more resource-intensive training configurations. Third, leading models increasingly rely on massive synthetic datasets comprising hundreds of millions of generated images~\cite{new_1_bartoldson2024adversarial, new_4_peng2023robust}. When the volume of synthetic training data grows sufficiently large, the boundary between training and test distributions becomes blurred, potentially introducing systematic optimistic bias into robustness estimates~\cite{Rethinking_RobustBench}. Taken together, these confounding factors make it exceedingly difficult to identify which algorithmic contributions truly advance the state of the art.

To address this crisis, we introduce the \textbf{FastAT Benchmark}, a rigorously controlled evaluation framework designed specifically for the fair comparison of FastAT methods. The benchmark is built upon three core design principles, each directly targeting one of the identified confounding factors:

\begin{enumerate}
    \item \textbf{Unified Architecture.} All methods are evaluated using identical network architectures, eliminating performance differences that arise from architectural advantages rather than training algorithms.
    \item \textbf{Standardized Training Settings.} Consistent training schedules, optimizers, learning rate policies, and data augmentation strategies are enforced across all methods, preventing the experimental setup from favoring any particular approach.
    \item \textbf{No External Data.} The use of any additional or synthetic data beyond the original benchmark training set is strictly prohibited, ensuring that all observed performance gains are attributable solely to the training algorithm itself.
\end{enumerate}

Building on these principles, the FastAT Benchmark provides a unified codebase that implements over twenty representative FastAT methods, enabling direct and reproducible comparisons across different algorithmic innovations. The benchmark further employs a dual-metric evaluation framework that captures both robustness and efficiency. Robustness is assessed through accuracy under a suite of strong adversarial attacks, including PGD with varying iteration counts \cite{pgd}, AutoAttack (AA) \cite{auto_attack}, and CR Attack \cite{cr_attack}. Efficiency is quantified by total GPU training time and peak memory footprint. 

The main contributions of this work are summarized as follows:

\begin{itemize}
    \item We identify the comparability crisis in FastAT research, highlighting confounding factors such as architectural heterogeneity, inconsistent training configurations, and external data reliance that undermine fair evaluation.
    
    \item We propose the FastAT Benchmark, a controlled evaluation framework that enforces unified architectures, standardized training settings, and strict prohibition of external data. The benchmark implements over twenty representative FastAT methods within a unified codebase, ensuring reproducibility and enabling direct comparison across different algorithmic innovations.
    
    \item We conduct comprehensive experiments on CIFAR-10, CIFAR-100, and Tiny-ImageNet, establishing reliable baseline measurements for the research community. The complete benchmark, including source code, configuration files, and full experimental results, is publicly released to support transparent and fair evaluation of future FastAT methods.
\end{itemize}

The remainder of this paper is organized as follows. Section~\ref{sec:related_work} reviews related work on adversarial training and existing evaluation frameworks. Section~\ref{sec:benchmark} presents the design and structure of the FastAT Benchmark. Section~\ref{sec:results} reports comprehensive experimental results. Finally, Section~\ref{sec:conclusion} concludes the paper.

\section{Related Work}
\label{sec:related_work}

\subsection{Adversarial Training}

Adversarial training remains one of the most reliable defenses against adversarial examples. Goodfellow et al.~\cite{fgsm} introduced FGSM-AT, the first adversarial training method, which generates perturbations using a single gradient step. Building on this foundation, Madry et al.~\cite{pgd} proposed PGD-AT, which employs multi-step projected gradient descent to construct stronger adversarial examples and has since become the de facto standard for adversarial training. Despite its strong robustness guarantees, PGD-AT requires multiple forward and backward passes at each training iteration, resulting in computational overhead that scales poorly to large datasets and architectures. This cost has motivated research into fast adversarial training methods that seek to match the robustness of PGD-AT at a significantly lower training budget.

\subsection{Fast Adversarial Training}

A central obstacle in fast adversarial training is catastrophic overfitting (CO)~\cite{fgsm_rs}, a phenomenon in which a model trained with single-step adversarial examples abruptly collapses to near-zero adversarial robustness during training, even as its clean accuracy remains high. Addressing CO has driven much of the algorithmic innovation in this field, and proposed solutions can be organized into several broad categories.

One line of work focuses on perturbation initialization. FGSM-RS~\cite{fgsm_rs} introduces random start initialization, which adds a uniform random perturbation before the gradient step and was among the first methods to successfully mitigate CO. N-FGSM~\cite{n_fgsm} extends this idea by sampling the uniform random initialization noise from a wider range beyond the perturbation budget $\varepsilon$ and correspondingly adjusting the FGSM step size. ZeroGrad~\cite{zero_grad} proposes a costless remedy that suppresses catastrophic gradient masking without additional computation. A second line of work uses informed or structured initialization. FGSM-PGI~\cite{fgsm_pgi} initializes perturbations from prior gradient information accumulated across training iterations, while FGSM-UAP, FGSM-CUAP, and FGSM-FUAP~\cite{fgsm_uap} leverage universal adversarial perturbations as a more expressive starting point. SSAT~\cite{ssat} contributes a theoretical analysis of CO and uses these insights to guide its training procedure.

Regularization-based approaches offer another perspective on stabilizing fast training. GradAlign~\cite{grad_align} directly penalizes the misalignment between input-space gradients at neighboring points, improving optimization stability. GAT~\cite{gat} incorporates guided adversarial examples to further regularize the loss landscape. Complementing these gradient-level methods, output consistency regularization approaches penalize inconsistent model predictions across the input neighborhood. NuAT~\cite{nu_at} enforces consistency under noise augmentation, AAER~\cite{aaer} regularizes the treatment of abnormal adversarial examples, and ELLE~\cite{elle} encourages local linearity of the loss surface.

Beyond single-step methods, FreeAT~\cite{free_at} pursues efficiency through a different mechanism: replaying each mini-batch multiple times while reusing gradient information. Several hybrid methods further combine multiple ideas. FGSM-MEP-CS and FGSM-RS-CS~\cite{fgsm_mep_cs_fgsm_rs_cs} improve convergence smoothness by blending perturbation strategies, FGSM-PCO~\cite{fgsm_pco} frames CO prevention as a bi-level optimization problem, and LIET~\cite{liet} mitigates CO by removing label information during training.

\subsection{Adversarial Robustness Benchmarks}

Standardized benchmarks have been essential for measuring progress in adversarial robustness. RobustBench~\cite{robustbench} is the most widely adopted framework in this area, providing a curated leaderboard evaluated with AA~\cite{auto_attack} under consistent threat models. It has significantly improved reproducibility and comparability across the robustness literature.

RobustBench, however, was not designed with the specific constraints of FastAT in mind. As discussed in Section~\ref{sec:introduction}, its leaderboard includes entries that differ substantially in model architecture, training budget, and reliance on large synthetic datasets, making it difficult to isolate the contribution of the training algorithm itself. Recent work has further shown that training on massive synthetic datasets can introduce optimistic bias into robustness estimates~\cite{Rethinking_RobustBench}. These shortcomings motivate a dedicated evaluation framework for FastAT methods, one that enforces controlled experimental conditions so that observed performance differences can be attributed reliably to algorithmic innovation rather than to confounding factors.

\section{The FastAT Benchmark}
\label{sec:benchmark}

The FastAT Benchmark is a controlled evaluation framework designed to enable fair and reproducible comparison across FastAT methods. As discussed in Section~\ref{sec:introduction}, existing evaluation practices introduce several confounding factors (architectural heterogeneity, inconsistent training configurations, and reliance on external data) that make it difficult to attribute performance differences to algorithmic choices alone. The benchmark addresses these issues by enforcing a set of strict experimental constraints, providing a unified codebase that implements over twenty representative methods, and evaluating all methods under a consistent and comprehensive protocol. The following subsections describe the core design principles, the methods covered, the implementation approach, the evaluation protocol, and the experimental configuration.

\subsection{Design Principles and Controlled Constraints}
\label{sec:design_principles}

The benchmark is built on three core design principles, each targeting one of the confounding factors identified in Section~\ref{sec:introduction}.

\textbf{Unified Architecture.} All methods are evaluated on identical network architectures. This constraint eliminates the possibility that reported performance improvements stem from architectural advantages rather than from the training algorithm itself.

\textbf{Standardized Training Settings.} Consistent training schedules, optimizers, learning rate policies, and data augmentation strategies are applied uniformly across all methods. By holding these factors constant, the benchmark ensures that no method benefits from a more favorable training configuration.

\textbf{No External Data.} The use of any additional or synthetic data beyond the original benchmark training set is strictly prohibited. This restriction ensures that all observed performance differences can be attributed solely to the learning algorithm, rather than to the scale or quality of training data.

Together, these three principles create a controlled experimental environment in which the algorithmic contribution of each method can be evaluated in isolation from external confounding variables.

\subsection{Supported Methods}
\label{sec:supported_methods}

The benchmark covers over twenty representative FastAT methods drawn from all major algorithmic categories in the literature. This breadth ensures that the benchmark is broadly representative of the current state of the field and allows for direct comparison across fundamentally different approaches to fast adversarial training.

\begin{itemize}
    \item \textbf{Baseline Approaches}: FGSM-AT~\cite{fgsm}, PGD-AT~\cite{pgd}, PGD-AT with weight averaging~\cite{wa}
    \item \textbf{Randomized Perturbation Initialization}: FGSM-RS~\cite{fgsm_rs}, N-FGSM~\cite{n_fgsm}, ZeroGrad~\cite{zero_grad}
    \item \textbf{Informed Initialization}: FGSM-PGI~\cite{fgsm_pgi}, FGSM-UAP, FGSM-CUAP, FGSM-FUAP~\cite{fgsm_uap}, SSAT~\cite{ssat}
    \item \textbf{Multi-step Approximation}: FreeAT~\cite{free_at}
    \item \textbf{Gradient Regularization}: GradAlign~\cite{grad_align}, GAT~\cite{gat}, ELLE~\cite{elle}
    \item \textbf{Output Consistency Regularization}: NuAT~\cite{nu_at}, AAER~\cite{aaer}
    \item \textbf{Hybrid and Data Augmentation Methods}: FGSM-PCO~\cite{fgsm_pco}, FGSM-MEP-CS, FGSM-RS-CS~\cite{fgsm_mep_cs_fgsm_rs_cs}, LIET~\cite{liet}
\end{itemize}

\subsection{Unified Implementation}
\label{sec:unified_impl}

A unified codebase is central to both the reproducibility and the fairness of the benchmark. All methods are \textit{re-implemented} within a shared framework that provides a common interface for data loading, model initialization, training loops, and evaluation protocols. This design prevents subtle implementation differences from inadvertently influencing results and makes it straightforward to incorporate new methods in the future.

The training workflow follows six sequential phases, applied consistently across all evaluated methods:

\begin{enumerate}
    \item \textbf{Load Configuration}: Load the shared \texttt{common.yaml} and the method-specific \texttt{method.yaml} configuration files.
    \item \textbf{Initialization}: Initialize data loaders, the model, and the optimizer according to the loaded configuration.
    \item \textbf{Method Selection}: Select and instantiate the specific FastAT method to be evaluated.
    \item \textbf{Training}: Execute the training loop with periodic validation to monitor convergence.
    \item \textbf{Evaluation}: Perform final evaluation using PGD, AA, and CR Attack.
    \item \textbf{Logging}: Record all experimental results and metrics for subsequent analysis.
\end{enumerate}

\subsection{Evaluation Protocol}
\label{sec:eval_protocol}

Because the central goal of FastAT is to achieve robustness efficiently, evaluating these methods requires assessing both their defensive effectiveness and their computational cost. The benchmark therefore employs a dual-metric evaluation framework that captures both dimensions.

The first dimension is robustness, measured as model accuracy under a suite of strong adversarial attacks. PGD attacks with 10, 20, and 50 iterations~\cite{pgd} assess robustness across a range of attack strengths. AA~\cite{auto_attack}, a widely adopted parameter-free ensemble attack, serves as a rigorous and reliable standard for robustness evaluation. CR Attack~\cite{cr_attack} is included as a supplementary measure. Together, these attacks provide a comprehensive and multi-faceted assessment of each method's defensive capabilities.

The second dimension is computational cost, quantified by total GPU training time and peak GPU memory footprint. Measuring this dimension explicitly rewards methods that are not only robust but also practical for real-world deployment. All metrics are reported as the mean and standard deviation across three independent runs with different random seeds, providing a reliable estimate of expected performance and its variability.

\subsection{Experimental Setup}
\label{sec:experimental_setup}

\subsubsection{Datasets and Models}

The benchmark evaluates all methods on three standard image classification datasets: CIFAR-10~\cite{cifar10}, CIFAR-100~\cite{cifar10}, and Tiny-ImageNet~\cite{deng2009imagenet}. For the two CIFAR datasets, the ResNet-18 architecture~\cite{resnet} is used. For Tiny-ImageNet, PreActResNet-18~\cite{preresnet} is adopted to maintain comparability with prior work. Each experiment is repeated three times with different random seeds (0, 1, 2), and the mean of these runs is reported as the final result.

\subsubsection{Training Configuration}

All models are trained using SGD with an initial learning rate of 0.1, a momentum of 0.9, and a weight decay of $5 \times 10^{-4}$. Training proceeds for 100 epochs with a OneCycleLR learning rate scheduler and a batch size of 128.

Standard data augmentation consists of 4-pixel padding followed by random cropping and horizontal flipping. Two additional techniques are applied uniformly across all methods: weight averaging~\cite{wa} with a decay rate of 0.9995, and label smoothing~\cite{label_smoothing} with a smoothing factor of 0.4. Because these techniques are applied consistently to every method, they function as orthogonal improvements and do not introduce any comparative advantage.

\subsubsection{Model Selection}

To select the final model for evaluation, a held-out validation set is reserved from the training data prior to any training. The validation set sizes are 1,000 images for CIFAR-10, 1,000 images for CIFAR-100, and 2,000 images for Tiny-ImageNet. The checkpoint achieving the highest PGD-10 accuracy on its corresponding validation set is retained as the final model. This selection strategy is applied uniformly across all methods to ensure a consistent and fair comparison.

\section{Experimental Results}
\label{sec:results}

We conduct comprehensive experiments on CIFAR-10, CIFAR-100, and Tiny-ImageNet to evaluate over twenty representative FastAT methods under strictly controlled settings. The following subsections analyze the results from four complementary perspectives: overall robustness and efficiency, training stability, multi-dimensional comparisons, and Pareto-optimal trade-off analysis.

\subsection{Robustness and Efficiency Performance}
\label{subsec:main_results}

Tables~\ref{tab:cifar10_results}, \ref{tab:cifar100_results}, and \ref{tab:tiny_imagenet_results} report clean accuracy, robust accuracy, training time, and peak GPU memory for all methods on CIFAR-10, CIFAR-100, and Tiny-ImageNet, respectively. Three consistent trends emerge across all benchmarks.

\begin{table*}[t]
\centering
\caption{Performance of FastAT methods on Cifar10. Results are reported as mean $\pm$ standard deviation across three runs.}
\label{tab:cifar10_results}
\resizebox{\textwidth}{!}{%
\begin{tabular}{lcccccccc}
\toprule
Method & Clean (\%) & PGD-10 (\%) & PGD-20 (\%) & PGD-50 (\%) & AA (\%) & CR (\%) & Time (s) & Mem (GB)\\
\midrule
ELLE & 77.78 $\pm$ 8.82 & 50.74 $\pm$ 3.46 & 48.89 $\pm$ 5.98 & 47.19 $\pm$ 8.82 & 32.46 $\pm$ 25.27 & 39.77 $\pm$ 12.77 & 2942.99 & 2.50 \\
FGSM-AT & 69.17 $\pm$ 0.95 & 42.99 $\pm$ 0.33 & 42.59 $\pm$ 0.28 & 42.49 $\pm$ 0.27 & 37.28 $\pm$ 0.30 & 37.27 $\pm$ 0.27 & 1996.07 & 1.39 \\
FGSM-CUAP & 80.76 $\pm$ 2.10 & 56.30 $\pm$ 0.02 & 55.69 $\pm$ 0.15 & 55.58 $\pm$ 0.22 & 49.55 $\pm$ 0.22 & 49.54 $\pm$ 0.22 & 2706.19 & 1.40 \\
FGSM-FUAP & 80.64 $\pm$ 1.28 & 56.35 $\pm$ 0.31 & 55.85 $\pm$ 0.33 & 55.75 $\pm$ 0.38 & 49.73 $\pm$ 0.12 & 49.71 $\pm$ 0.12 & 2927.89 & 1.40 \\
FGSM-MEP-CS & 83.55 $\pm$ 3.14 & 51.11 $\pm$ 0.54 & 50.09 $\pm$ 0.32 & 49.80 $\pm$ 0.36 & 45.35 $\pm$ 0.81 & 45.37 $\pm$ 0.78 & 2141.50 & 3.66 \\
FGSM-PCO & 84.69 $\pm$ 0.32 & 55.04 $\pm$ 0.25 & 54.36 $\pm$ 0.18 & 54.18 $\pm$ 0.21 & 48.52 $\pm$ 0.32 & 48.48 $\pm$ 0.34 & 2997.04 & 2.54 \\
FGSM-PGI & 79.72 $\pm$ 0.63 & 56.43 $\pm$ 0.18 & 55.91 $\pm$ 0.18 & 55.82 $\pm$ 0.21 & 49.89 $\pm$ 0.08 & 49.89 $\pm$ 0.12 & 2646.65 & 2.54 \\
FGSM-RS & 77.66 $\pm$ 4.65 & 49.15 $\pm$ 5.09 & 48.38 $\pm$ 5.28 & 48.27 $\pm$ 5.33 & 43.09 $\pm$ 5.33 & 43.14 $\pm$ 5.29 & 1902.17 & 1.39 \\
FGSM-RS-CS & 80.54 $\pm$ 2.06 & 53.99 $\pm$ 0.23 & 53.37 $\pm$ 0.22 & 53.12 $\pm$ 0.29 & 48.64 $\pm$ 0.38 & 48.70 $\pm$ 0.34 & 1824.07 & 1.95 \\
FGSM-UAP & 80.61 $\pm$ 0.48 & 56.49 $\pm$ 0.17 & 55.90 $\pm$ 0.07 & 55.77 $\pm$ 0.13 & 49.44 $\pm$ 0.14 & 49.37 $\pm$ 0.15 & 2551.80 & 1.40 \\
FREE-AT & 75.20 $\pm$ 0.43 & 49.32 $\pm$ 0.22 & 48.92 $\pm$ 0.29 & 48.83 $\pm$ 0.26 & 44.34 $\pm$ 0.17 & 44.32 $\pm$ 0.24 & 1978.59 & 1.40 \\
GAT & 85.10 $\pm$ 0.13 & 54.91 $\pm$ 0.06 & 53.98 $\pm$ 0.09 & 53.81 $\pm$ 0.14 & 49.15 $\pm$ 0.15 & 49.15 $\pm$ 0.17 & 2875.82 & 1.90 \\
GRAD-ALIGN & 79.39 $\pm$ 1.97 & 53.57 $\pm$ 0.44 & 53.02 $\pm$ 0.31 & 52.87 $\pm$ 0.29 & 47.99 $\pm$ 0.62 & 48.02 $\pm$ 0.61 & 5612.39 & 2.01 \\
LIET & 80.68 $\pm$ 0.51 & 56.69 $\pm$ 0.18 & 56.12 $\pm$ 0.19 & 55.97 $\pm$ 0.19 & 49.99 $\pm$ 0.15 & 49.93 $\pm$ 0.15 & 2605.29 & 1.40 \\
N-AAER & 77.67 $\pm$ 0.48 & 52.84 $\pm$ 0.16 & 52.44 $\pm$ 0.18 & 52.37 $\pm$ 0.15 & 47.38 $\pm$ 0.18 & 47.48 $\pm$ 0.20 & 1863.55 & 1.39 \\
N-FGSM & 78.59 $\pm$ 2.34 & 53.64 $\pm$ 0.25 & 53.13 $\pm$ 0.27 & 53.07 $\pm$ 0.28 & 47.77 $\pm$ 0.41 & 47.84 $\pm$ 0.41 & 1940.75 & 1.39 \\
NU-AT & 80.98 $\pm$ 0.91 & 55.33 $\pm$ 0.14 & 54.78 $\pm$ 0.20 & 54.63 $\pm$ 0.25 & 50.10 $\pm$ 0.18 & 50.11 $\pm$ 0.19 & 2936.85 & 1.90 \\
PGD-AT & 85.35 $\pm$ 0.14 & 48.35 $\pm$ 0.18 & 46.74 $\pm$ 0.15 & 46.31 $\pm$ 0.10 & 44.33 $\pm$ 0.08 & 44.50 $\pm$ 0.13 & 6953.97 & 1.30 \\
PGD-AT-WA & 82.36 $\pm$ 0.32 & 55.05 $\pm$ 0.34 & 54.28 $\pm$ 0.20 & 54.17 $\pm$ 0.17 & 50.49 $\pm$ 0.15 & 50.51 $\pm$ 0.16 & 7026.80 & 1.39 \\
SSAT & 84.80 $\pm$ 2.24 & 52.40 $\pm$ 0.45 & 51.70 $\pm$ 0.53 & 51.46 $\pm$ 0.57 & 38.33 $\pm$ 1.55 & 38.32 $\pm$ 1.53 & 2120.88 & 1.40 \\
ZERO-GRAD & 75.85 $\pm$ 3.07 & 51.59 $\pm$ 2.06 & 51.12 $\pm$ 1.97 & 51.06 $\pm$ 1.98 & 45.94 $\pm$ 2.19 & 46.04 $\pm$ 2.20 & 1924.46 & 1.39 \\
\bottomrule
\end{tabular}}
\end{table*}

\begin{table*}[!t]
\centering
\caption{Performance of FastAT methods on Cifar100. Results are reported as mean $\pm$ standard deviation across three runs.}
\label{tab:cifar100_results}
\resizebox{\textwidth}{!}{%
\begin{tabular}{lcccccccc}
\toprule
Method & Clean (\%) & PGD-10 (\%) & PGD-20 (\%) & PGD-50 (\%) & AA (\%) & CR (\%) & Time (s) & Mem (GB)\\
\midrule
ELLE & 47.48 $\pm$ 3.45 & 26.68 $\pm$ 2.45 & 26.40 $\pm$ 2.53 & 26.34 $\pm$ 2.53 & 21.21 $\pm$ 2.37 & 21.23 $\pm$ 2.37 & 2934.95 & 2.51 \\
FGSM-AT & 39.50 $\pm$ 0.77 & 21.42 $\pm$ 0.55 & 21.25 $\pm$ 0.52 & 21.18 $\pm$ 0.54 & 16.62 $\pm$ 0.54 & 16.58 $\pm$ 0.50 & 2072.48 & 1.39 \\
FGSM-CUAP & 55.12 $\pm$ 1.80 & 31.65 $\pm$ 0.26 & 31.43 $\pm$ 0.28 & 31.40 $\pm$ 0.31 & 25.85 $\pm$ 0.50 & 25.83 $\pm$ 0.49 & 2865.16 & 1.40 \\
FGSM-FUAP & 56.29 $\pm$ 1.13 & 31.76 $\pm$ 0.31 & 31.42 $\pm$ 0.51 & 31.36 $\pm$ 0.46 & 25.95 $\pm$ 0.08 & 25.94 $\pm$ 0.05 & 3009.01 & 1.41 \\
FGSM-MEP-CS & 59.60 $\pm$ 1.88 & 31.42 $\pm$ 0.12 & 30.99 $\pm$ 0.03 & 30.86 $\pm$ 0.11 & 24.86 $\pm$ 0.28 & 24.80 $\pm$ 0.26 & 2113.57 & 3.66 \\
FGSM-PCO & 64.45 $\pm$ 1.08 & 26.32 $\pm$ 0.42 & 25.75 $\pm$ 0.44 & 25.69 $\pm$ 0.28 & 19.97 $\pm$ 0.43 & 19.91 $\pm$ 0.39 & 2828.23 & 2.54 \\
FGSM-PGI & 56.61 $\pm$ 0.34 & 32.18 $\pm$ 0.39 & 31.88 $\pm$ 0.32 & 31.85 $\pm$ 0.23 & 26.29 $\pm$ 0.16 & 26.31 $\pm$ 0.12 & 2656.53 & 2.52 \\
FGSM-RS & 49.00 $\pm$ 3.64 & 27.50 $\pm$ 2.54 & 27.36 $\pm$ 2.49 & 27.32 $\pm$ 2.51 & 21.42 $\pm$ 2.24 & 21.46 $\pm$ 2.24 & 2065.97 & 1.39 \\
FGSM-RS-CS & 54.79 $\pm$ 1.07 & 32.42 $\pm$ 0.29 & 32.11 $\pm$ 0.25 & 32.08 $\pm$ 0.29 & 26.20 $\pm$ 0.26 & 26.20 $\pm$ 0.26 & 1878.41 & 1.95 \\
FGSM-UAP & 57.84 $\pm$ 0.28 & 31.43 $\pm$ 0.24 & 31.19 $\pm$ 0.25 & 31.07 $\pm$ 0.31 & 25.84 $\pm$ 0.24 & 25.82 $\pm$ 0.23 & 2662.54 & 1.40 \\
FREE-AT & 48.98 $\pm$ 0.19 & 27.39 $\pm$ 0.31 & 27.22 $\pm$ 0.34 & 27.19 $\pm$ 0.34 & 21.76 $\pm$ 0.25 & 21.74 $\pm$ 0.27 & 1980.92 & 1.39 \\
GAT & 65.52 $\pm$ 0.34 & 27.66 $\pm$ 0.16 & 26.80 $\pm$ 0.12 & 26.67 $\pm$ 0.16 & 22.06 $\pm$ 0.08 & 21.97 $\pm$ 0.07 & 2945.59 & 1.90 \\
GRAD-ALIGN & 54.61 $\pm$ 0.75 & 31.21 $\pm$ 0.05 & 30.98 $\pm$ 0.05 & 30.92 $\pm$ 0.02 & 25.37 $\pm$ 0.14 & 25.39 $\pm$ 0.15 & 5613.01 & 2.01 \\
LIET & 50.47 $\pm$ 0.69 & 32.64 $\pm$ 0.53 & 32.46 $\pm$ 0.46 & 32.44 $\pm$ 0.51 & 26.64 $\pm$ 0.13 & 26.63 $\pm$ 0.14 & 2343.70 & 1.40 \\
N-AAER & 51.95 $\pm$ 0.75 & 27.37 $\pm$ 1.38 & 27.04 $\pm$ 1.38 & 26.95 $\pm$ 1.43 & 22.74 $\pm$ 1.17 & 22.81 $\pm$ 1.18 & 1731.43 & 1.39 \\
N-FGSM & 53.22 $\pm$ 2.73 & 30.58 $\pm$ 0.25 & 30.40 $\pm$ 0.22 & 30.35 $\pm$ 0.25 & 24.86 $\pm$ 0.46 & 24.89 $\pm$ 0.49 & 2063.26 & 1.39 \\
NU-AT & 52.63 $\pm$ 14.39 & 25.37 $\pm$ 1.81 & 23.04 $\pm$ 0.92 & 21.63 $\pm$ 1.70 & 14.41 $\pm$ 3.29 & 17.66 $\pm$ 0.42 & 3092.07 & 1.90 \\
PGD-AT & 59.11 $\pm$ 0.39 & 26.68 $\pm$ 0.09 & 25.93 $\pm$ 0.04 & 25.82 $\pm$ 0.09 & 21.44 $\pm$ 0.23 & 20.81 $\pm$ 0.30 & 6724.17 & 1.31 \\
PGD-AT-WA & 57.62 $\pm$ 0.34 & 32.31 $\pm$ 0.16 & 31.99 $\pm$ 0.16 & 31.90 $\pm$ 0.14 & 26.90 $\pm$ 0.17 & 26.87 $\pm$ 0.17 & 7015.43 & 1.39 \\
SSAT & 66.79 $\pm$ 1.11 & 24.88 $\pm$ 0.52 & 23.49 $\pm$ 0.54 & 23.10 $\pm$ 0.56 & 18.83 $\pm$ 0.73 & 18.79 $\pm$ 0.79 & 2134.76 & 1.40 \\
ZERO-GRAD & 57.25 $\pm$ 0.98 & 30.70 $\pm$ 0.26 & 30.23 $\pm$ 0.35 & 30.15 $\pm$ 0.32 & 25.00 $\pm$ 0.14 & 24.95 $\pm$ 0.14 & 2025.50 & 1.39 \\
\bottomrule
\end{tabular}}
\end{table*}

\begin{table*}[t]
\centering
\caption{Performance of FastAT methods on Tiny Imagenet. Results are reported as mean $\pm$ standard deviation across three runs.}
\label{tab:tiny_imagenet_results}
\resizebox{\textwidth}{!}{%
\begin{tabular}{lcccccccc}
\toprule
Method & Clean (\%) & PGD-10 (\%) & PGD-20 (\%) & PGD-50 (\%) & AA (\%) & CR (\%) & Time (s) & Mem (GB)\\
\midrule
ELLE & 43.70 $\pm$ 0.83 & 25.15 $\pm$ 0.12 & 24.95 $\pm$ 0.11 & 24.91 $\pm$ 0.12 & 18.81 $\pm$ 0.21 & 18.80 $\pm$ 0.21 & 19270.18 & 8.85 \\
FGSM-AT & 39.99 $\pm$ 7.05 & 21.14 $\pm$ 4.51 & 20.92 $\pm$ 4.43 & 20.92 $\pm$ 4.43 & 14.87 $\pm$ 3.77 & 14.84 $\pm$ 3.75 & 8109.43 & 4.56 \\
FGSM-CUAP & 47.22 $\pm$ 0.89 & 26.34 $\pm$ 0.38 & 26.13 $\pm$ 0.38 & 26.13 $\pm$ 0.38 & 19.19 $\pm$ 0.09 & 19.20 $\pm$ 0.11 & 12098.58 & 4.60 \\
FGSM-FUAP & 48.45 $\pm$ 1.25 & 26.32 $\pm$ 0.15 & 26.10 $\pm$ 0.17 & 26.09 $\pm$ 0.17 & 19.10 $\pm$ 0.18 & 19.09 $\pm$ 0.19 & 13111.93 & 4.60 \\
FGSM-MEP-CS & 49.76 $\pm$ 0.24 & 26.47 $\pm$ 0.25 & 26.20 $\pm$ 0.22 & 26.17 $\pm$ 0.25 & 18.23 $\pm$ 0.19 & 18.13 $\pm$ 0.16 & 10622.89 & 21.21 \\
FGSM-PCO & 59.67 $\pm$ 0.84 & 19.61 $\pm$ 0.15 & 18.93 $\pm$ 0.09 & 18.72 $\pm$ 0.09 & 11.24 $\pm$ 0.21 & 11.20 $\pm$ 0.23 & 15338.78 & 13.84 \\
FGSM-PGI & 47.89 $\pm$ 0.99 & 26.57 $\pm$ 0.03 & 26.32 $\pm$ 0.04 & 26.29 $\pm$ 0.07 & 19.55 $\pm$ 0.11 & 19.52 $\pm$ 0.12 & 11200.37 & 18.57 \\
FGSM-RS & 43.91 $\pm$ 3.34 & 23.06 $\pm$ 0.59 & 22.87 $\pm$ 0.64 & 22.85 $\pm$ 0.62 & 16.33 $\pm$ 0.67 & 16.26 $\pm$ 0.71 & 8128.66 & 4.56 \\
FGSM-RS-CS & 43.63 $\pm$ 0.55 & 25.90 $\pm$ 0.16 & 25.71 $\pm$ 0.19 & 25.71 $\pm$ 0.19 & 18.74 $\pm$ 0.11 & 18.69 $\pm$ 0.11 & 10000.28 & 6.64 \\
FGSM-UAP & 49.39 $\pm$ 0.36 & 26.75 $\pm$ 0.27 & 26.46 $\pm$ 0.29 & 26.37 $\pm$ 0.29 & 18.90 $\pm$ 0.28 & 18.96 $\pm$ 0.21 & 10994.11 & 4.58 \\
FREE-AT & 36.06 $\pm$ 0.33 & 19.96 $\pm$ 0.21 & 19.85 $\pm$ 0.22 & 19.81 $\pm$ 0.21 & 13.51 $\pm$ 0.12 & 13.51 $\pm$ 0.15 & 8293.47 & 4.57 \\
GAT & 57.33 $\pm$ 0.61 & 17.83 $\pm$ 0.20 & 17.06 $\pm$ 0.16 & 16.94 $\pm$ 0.19 & 11.33 $\pm$ 0.06 & 11.27 $\pm$ 0.06 & 14382.82 & 6.57 \\
GRAD-ALIGN & 45.43 $\pm$ 0.49 & 24.54 $\pm$ 0.31 & 24.35 $\pm$ 0.24 & 24.32 $\pm$ 0.24 & 17.87 $\pm$ 0.19 & 17.81 $\pm$ 0.22 & 35944.00 & 6.76 \\
LIET & 44.80 $\pm$ 0.08 & 26.64 $\pm$ 0.19 & 26.54 $\pm$ 0.17 & 26.52 $\pm$ 0.17 & 19.88 $\pm$ 0.09 & 19.86 $\pm$ 0.07 & 11463.39 & 4.61 \\
N-FGSM & 47.50 $\pm$ 1.55 & 25.52 $\pm$ 0.14 & 25.28 $\pm$ 0.22 & 25.22 $\pm$ 0.18 & 18.79 $\pm$ 0.08 & 18.76 $\pm$ 0.06 & 8163.30 & 4.56 \\
NU-AT & 26.63 $\pm$ 3.10 & 15.15 $\pm$ 0.54 & 15.08 $\pm$ 0.52 & 15.08 $\pm$ 0.55 & 10.20 $\pm$ 0.44 & 10.19 $\pm$ 0.44 & 15308.53 & 6.58 \\
PGD-AT & 51.37 $\pm$ 0.19 & 23.36 $\pm$ 0.21 & 22.90 $\pm$ 0.25 & 22.78 $\pm$ 0.18 & 17.04 $\pm$ 0.08 & 16.84 $\pm$ 0.12 & 28575.32 & 4.47 \\
PGD-AT-WA & 48.07 $\pm$ 0.97 & 26.38 $\pm$ 0.22 & 26.19 $\pm$ 0.21 & 26.15 $\pm$ 0.22 & 19.77 $\pm$ 0.16 & 19.75 $\pm$ 0.13 & 28572.77 & 4.56 \\
SSAT & 59.82 $\pm$ 0.66 & 18.75 $\pm$ 0.14 & 17.82 $\pm$ 0.16 & 17.55 $\pm$ 0.09 & 11.94 $\pm$ 0.12 & 11.92 $\pm$ 0.14 & 9458.99 & 4.58 \\
ZERO-GRAD & 46.14 $\pm$ 3.99 & 25.02 $\pm$ 1.14 & 24.81 $\pm$ 1.06 & 24.73 $\pm$ 1.00 & 18.15 $\pm$ 1.11 & 18.10 $\pm$ 1.14 & 8218.72 & 4.56 \\
\bottomrule
\end{tabular}}
\end{table*}

\paragraph{Clean-robust accuracy trade-off.}
A pronounced trade-off between clean accuracy and robust accuracy is evident across all three datasets. Methods such as SSAT, GAT, and FGSM-PCO consistently achieve the highest clean accuracies, yet their AA and CR scores are among the lowest. On CIFAR-100, for example, SSAT and GAT attain clean accuracies of 66.79\% and 65.52\%, respectively, while their AA scores fall to only 18.83\% and 22.06\%. This contrast is further amplified on Tiny-ImageNet, where SSAT (59.82\% clean) and FGSM-PCO (59.67\% clean) yield only 11.94\% and 11.24\% AA, respectively. These results illustrate the fundamental tension between optimizing for natural generalization and adversarial robustness.

\paragraph{Competitiveness of single-step methods.}
Contrary to the assumption that multi-step training inherently confers superior robustness, carefully designed single-step methods prove highly competitive with PGD-AT-WA across all benchmarks. On CIFAR-10, PGD-AT-WA achieves 50.49\% AA at a training cost of 7,027s, while LIET, FGSM-PGI, and NU-AT approach or exceed 50\% AA at roughly one-third to one-half the training time. This pattern holds on CIFAR-100 as well. On Tiny-ImageNet, PGD-AT-WA achieves 19.77\% AA at a cost exceeding 28,000s, a score that LIET surpasses at less than half the training time.

\paragraph{Scaling behavior across datasets.}
As task complexity increases from CIFAR-10 to CIFAR-100 and Tiny-ImageNet, absolute accuracy values decrease substantially, reflecting the growing difficulty of each classification task. The relative performance rankings among methods, however, are broadly preserved across datasets, suggesting that algorithmic advantages generalize across benchmark scales. That said, some methods that perform competitively on simpler benchmarks degrade more sharply on harder ones, a pattern discussed further in the context of training stability below.

\subsection{Training Stability}
\label{subsec:stability}

Training stability, measured by the standard deviation of accuracy across multiple runs, varies considerably among methods and tends to deteriorate as benchmark difficulty increases. On CIFAR-10, ELLE and FGSM-RS exhibit the highest AA variance, with standard deviations of 25.27 and 5.33, respectively, indicating susceptibility to catastrophic overfitting and sensitivity to random initialization. On CIFAR-100, NU-AT suffers a dramatic increase in instability, with standard deviations of 14.39 on clean accuracy and 3.29 on AA. This instability intensifies further on Tiny-ImageNet, where NU-AT yields severely degraded performance. FGSM-AT similarly exhibits large instability on Tiny-ImageNet, with standard deviations of 7.05 and 3.77 on clean and AA accuracy, respectively.

By contrast, methods that employ principled perturbation initialization and explicit regularization, such as FGSM-PGI and PGD-AT-WA, maintain consistently low variance across all three benchmarks. These results suggest that training stability is a distinct and practically important property that merits consideration alongside robustness and efficiency.

\subsection{Computational Efficiency and Memory Usage}
\label{subsec:efficiency}

Beyond robustness and stability, computational cost and memory consumption are critical factors for practical deployment. Among the evaluated methods, FGSM-RS-CS stands out as particularly resource-efficient, achieving competitive AA robustness (48.64\% on CIFAR-10, 26.20\% on CIFAR-100) with training times of approximately 1,824s and 1,878s, respectively. At the other extreme, GRAD-ALIGN incurs the longest training time on Tiny-ImageNet (35,944s) while delivering only moderate AA (17.87\%), illustrating that gradient-alignment-based regularization becomes computationally prohibitive at larger input resolutions.

Memory consumption also varies substantially across methods. FGSM-MEP-CS incurs consistently high GPU memory usage without commensurate robustness gains, reaching 3.66 GB on CIFAR-100 and 21.21 GB on Tiny-ImageNet. FGSM-PGI (18.57 GB) and FGSM-PCO (13.84 GB) also impose considerable memory overhead on Tiny-ImageNet, limiting their scalability to larger datasets and resolutions. These observations underscore the importance of jointly evaluating robustness and resource efficiency, rather than optimizing for accuracy alone.

\subsection{Multi-Dimensional Performance Comparison}
\label{subsec:radar}

\begin{figure*}[htbp]
    \centering
    \begin{subfigure}[b]{0.32\textwidth}
        \centering
        \includegraphics[width=\textwidth]{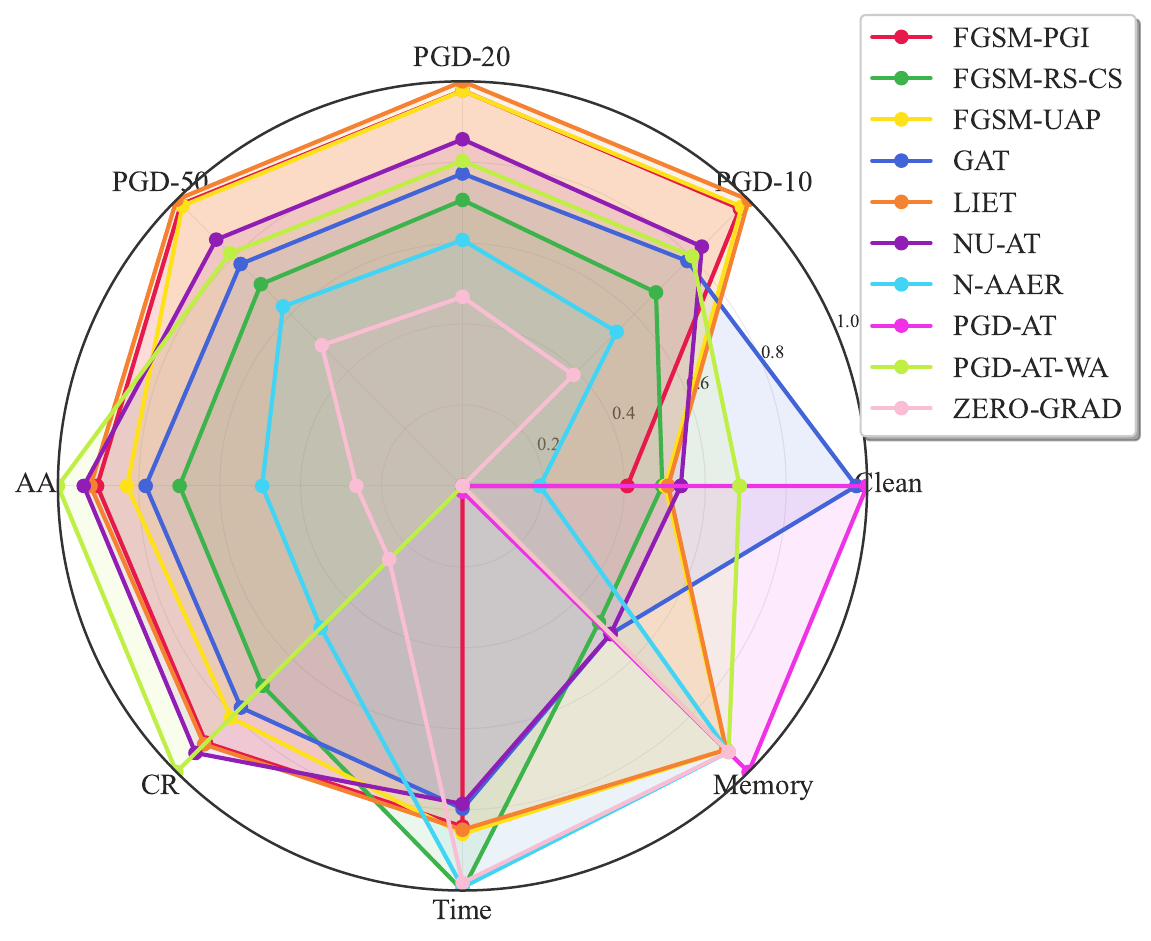}
        \caption{CIFAR-10}
        \label{fig:radar_cifar10}
    \end{subfigure}
    \hfill
    \begin{subfigure}[b]{0.32\textwidth}
        \centering
        \includegraphics[width=\textwidth]{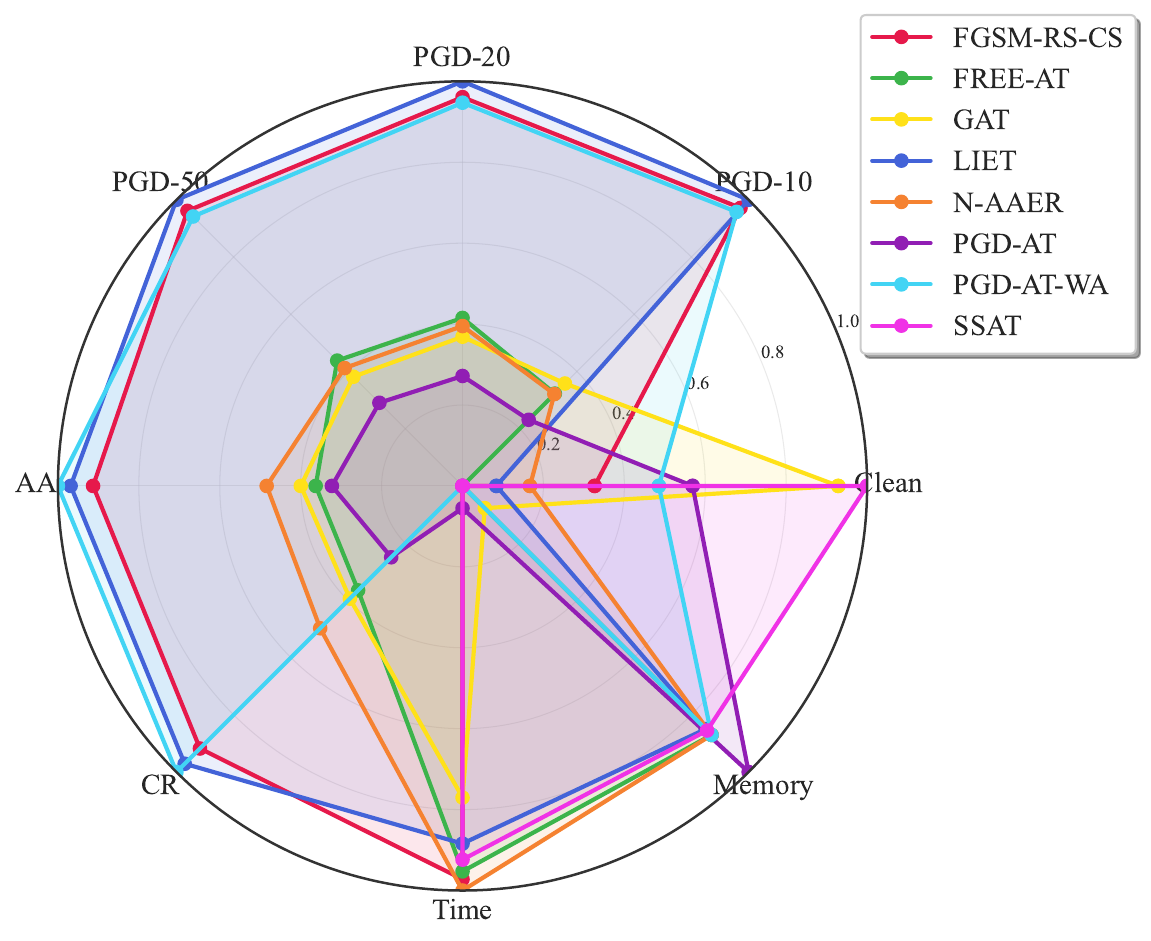}
        \caption{CIFAR-100}
        \label{fig:radar_cifar100}
    \end{subfigure}
    \hfill
    \begin{subfigure}[b]{0.32\textwidth}
        \centering
        \includegraphics[width=\textwidth]{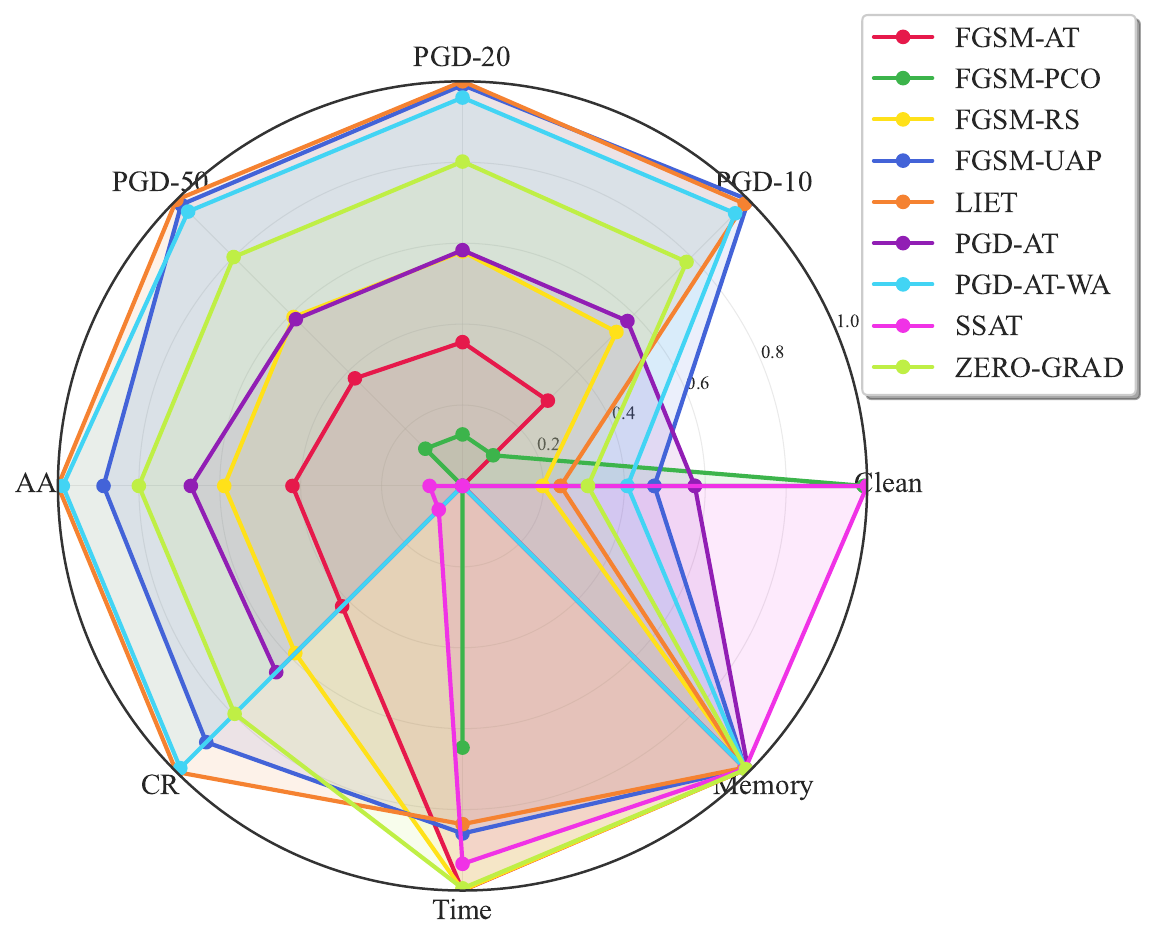}
        \caption{Tiny-ImageNet}
        \label{fig:radar_tinyimagenet}
    \end{subfigure}
    \caption{Radar charts of selected FastAT methods on CIFAR-10, CIFAR-100, and Tiny-ImageNet. Each axis represents a normalized performance metric (higher is better). The charts include the top-two methods per metric after deduplication.}
    \label{fig:radar_charts}
\end{figure*}

\begin{figure*}[t]
    \centering
    \begin{subfigure}[b]{0.32\textwidth}
        \includegraphics[width=\textwidth]{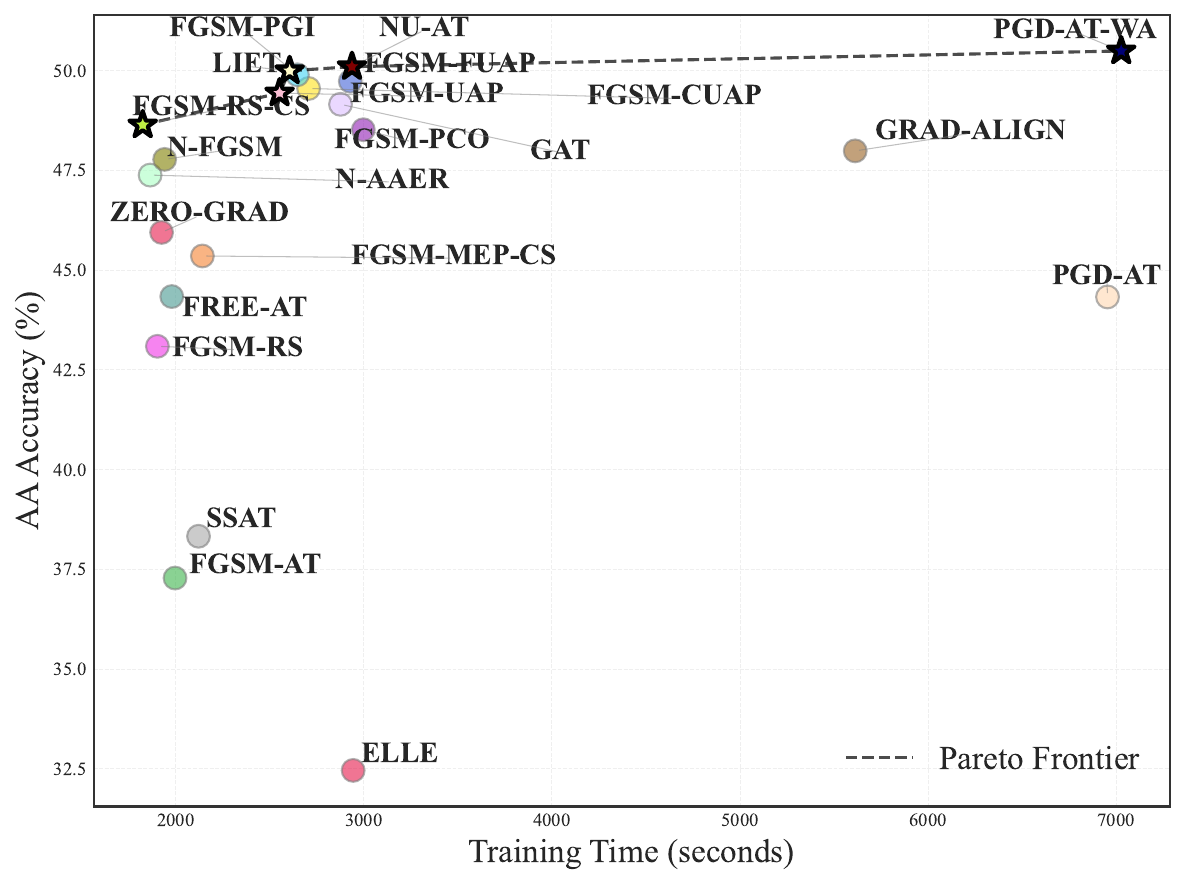}
        \caption{CIFAR-10}
        \label{fig:pareto_cifar10}
    \end{subfigure}
    \hfill
    \begin{subfigure}[b]{0.32\textwidth}
        \includegraphics[width=\textwidth]{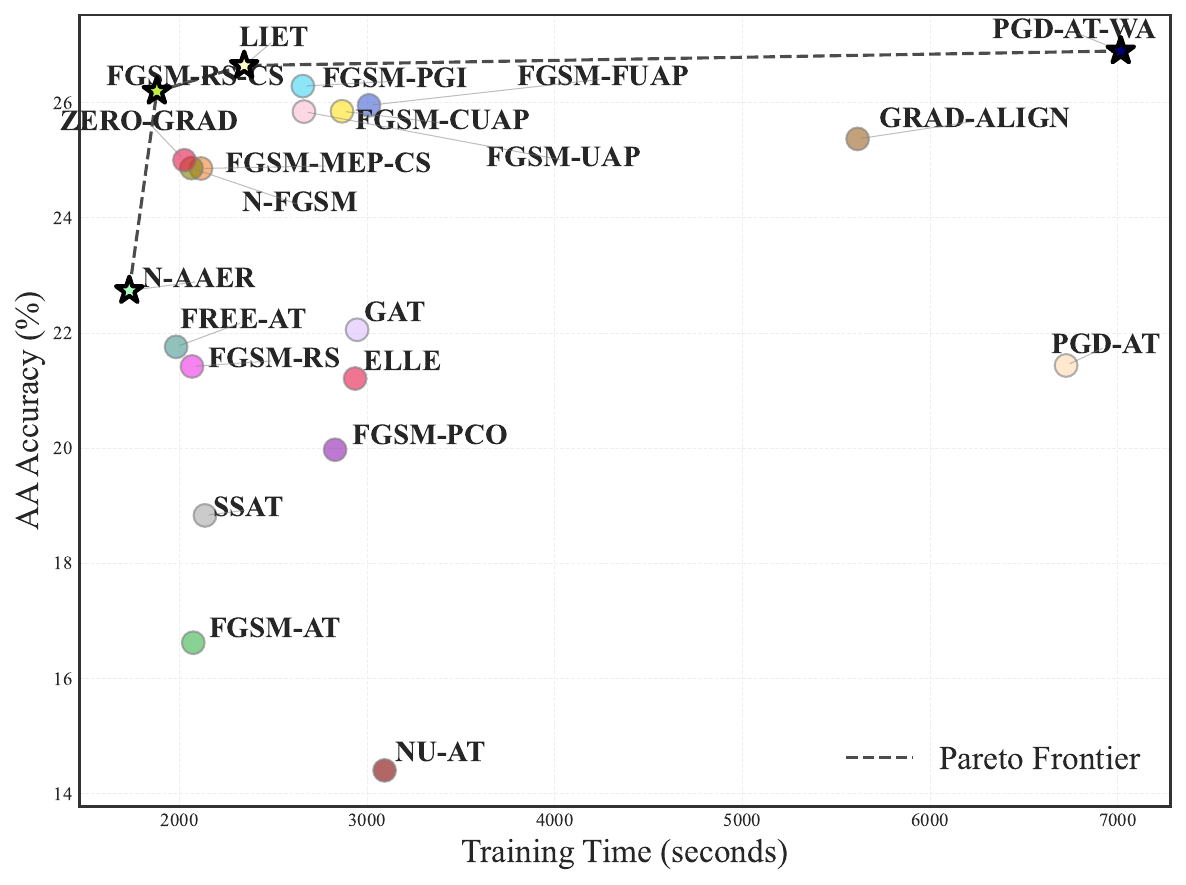}
        \caption{CIFAR-100}
        \label{fig:pareto_cifar100}
    \end{subfigure}
    \hfill
    \begin{subfigure}[b]{0.32\textwidth}
        \includegraphics[width=\textwidth]{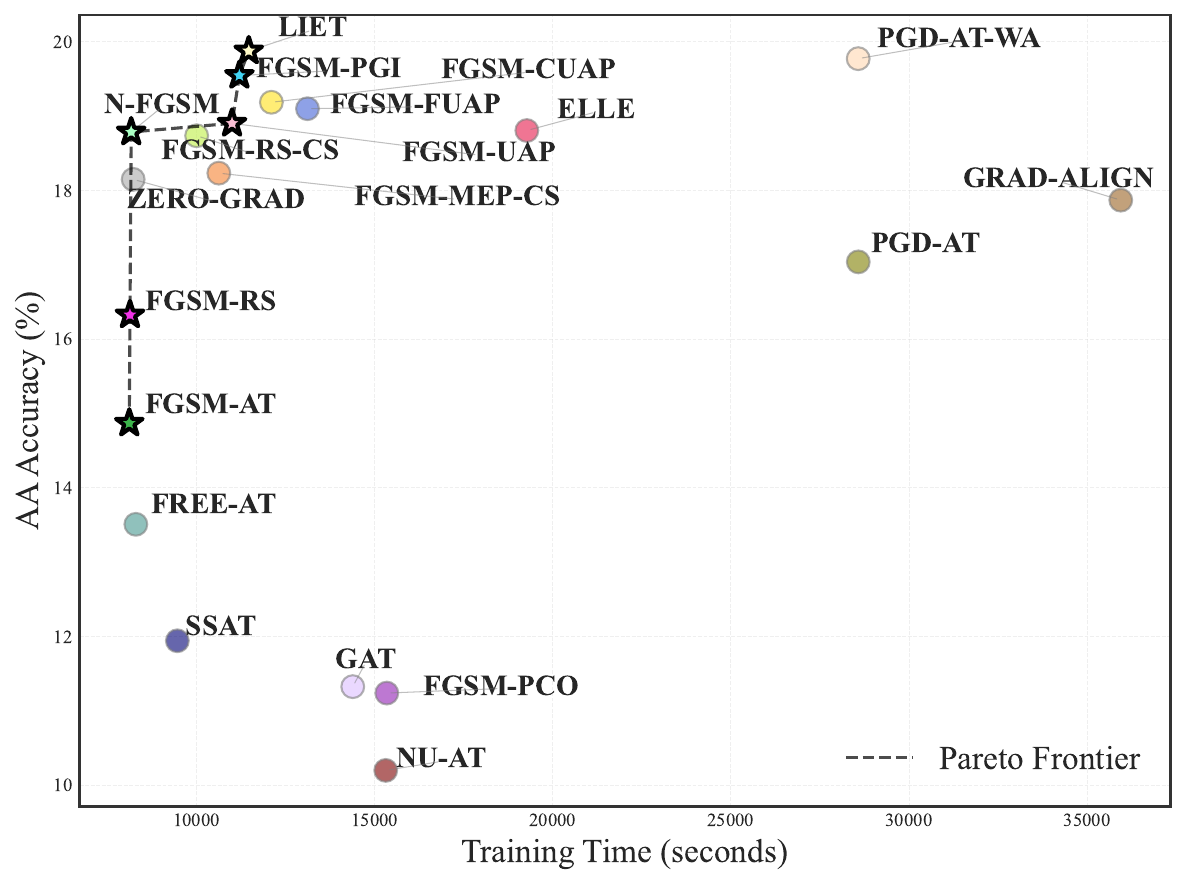}
        \caption{Tiny-ImageNet}
        \label{fig:pareto_tinyimagenet}
    \end{subfigure}
    \caption{Pareto frontiers illustrating the trade-off between training time and AutoAttack accuracy. Star markers denote Pareto-optimal methods achieving the best robustness-efficiency balance.}
    \label{fig:pareto_all}
\end{figure*}

To provide a holistic view of the trade-offs among FastAT methods, Fig.~\ref{fig:radar_charts} presents radar charts that simultaneously compare multiple performance dimensions: clean accuracy, robustness under PGD-10/20/50 attacks, AA, CR Attack, training time, and GPU memory usage. For each metric, the top-two performing methods are identified, with training time and memory usage inverted so that higher values correspond to better efficiency. The union of all selected methods is then visualized for each dataset.

These radar charts reveal several notable patterns. First, no single method dominates across all metrics, confirming that FastAT design involves inherent multi-dimensional trade-offs in which excelling along one dimension typically comes at a cost in another. Second, a non-trivial gap exists between PGD-based robustness and AA robustness. While LIET consistently achieves top-tier PGD-10/20/50 robustness across all three datasets, PGD-AT-WA demonstrates superior performance under AA and CR Attack. This discrepancy suggests that PGD-based evaluation alone may overestimate true robustness. Third, a subset of methods achieves relatively balanced performance across all dimensions. On CIFAR-10, FGSM-UAP and NU-AT maintain competitive robustness with reasonable efficiency; on Tiny-ImageNet, ZERO-GRAD demonstrates well-rounded performance across robustness, efficiency, and memory usage. Such balanced performers are particularly valuable for practitioners who cannot afford to sacrifice any single dimension.

\subsection{Robustness-Efficiency Trade-off Analysis}
\label{subsec:pareto}

To formally characterize the robustness-efficiency trade-off, Fig.~\ref{fig:pareto_all} visualizes the Pareto frontiers across all three datasets, identifying methods that achieve optimal trade-offs between AA accuracy and training time. On CIFAR-10, five methods form the Pareto frontier: FGSM-RS-CS achieves the fastest training time (1,824s) with 48.64\% AA accuracy, PGD-AT-WA attains the highest robustness (50.49\%) at significantly greater cost (7,027s), and NU-AT provides a well-balanced intermediate option. On CIFAR-100, four methods are Pareto-optimal: FGSM-RS-CS again demonstrates strong efficiency (1,878s, 26.20\% AA), while LIET achieves competitive robustness (26.64\% AA) at a moderate training time of 2,344s. On Tiny-ImageNet, six methods form the Pareto frontier, with N-FGSM emerging as a strong performer (18.79\% AA) at minimal overhead and LIET achieving the highest AA among all FastAT methods (19.88\%) at 11,463s.

Two key findings emerge from the Pareto analysis. First, FGSM-RS-CS provides excellent training efficiency with competitive robustness, particularly on smaller datasets, making it a strong candidate when the computational budget is the primary constraint. Second, the efficiency advantage of FastAT methods over PGD-AT-WA is most pronounced on CIFAR-10 and CIFAR-100, while this margin narrows on Tiny-ImageNet due to the increased absolute training times associated with higher input resolutions.

\subsection{Summary}
\label{subsec:results_summary}

Across all three benchmarks, the experimental results consistently support several overarching conclusions. Single-step FastAT methods can match or exceed the robustness of PGD-AT at a fraction of the computational cost, provided they incorporate principled perturbation initialization and regularization. No method achieves universally optimal performance across all evaluation dimensions, underscoring the inherently multi-objective nature of adversarial training. Training stability emerges as a critical and often overlooked property, particularly as dataset complexity increases. Finally, resource efficiency, encompassing both training time and memory consumption, must be considered alongside robustness when selecting a method for practical deployment. These findings collectively highlight the value of the controlled evaluation framework introduced by the FastAT Benchmark, which enables such nuanced, cross-method comparisons to be made reliably.

\section{Conclusion}
\label{sec:conclusion}

This work introduced the FastAT Benchmark, a controlled evaluation framework designed to address the comparability crisis in FastAT research. By enforcing three core principles (unified architecture, standardized training settings, and strict prohibition of external data) the benchmark isolates algorithmic contributions from confounding factors. Over twenty representative FastAT methods were implemented within a single codebase and evaluated on CIFAR-10, CIFAR-100, and Tiny-ImageNet using a dual-metric framework that captures both adversarial robustness and computational cost.

The experimental results reveal several consistent findings. Well-designed single-step methods can match or even surpass the robustness of PGD-AT at a substantially lower training cost. At the same time, no single method dominates across all evaluation dimensions, confirming that adversarial training is inherently a multi-objective problem. Training stability emerges as a critical and often underappreciated factor, particularly as dataset complexity increases.

We hope the FastAT Benchmark serves as a reliable foundation for future research, enabling fair and reproducible comparisons as new methods continue to emerge. The full benchmark, including source code, configuration files, and experimental results, is publicly released to support the community in this direction.

\bibliographystyle{IEEEtran}
\bibliography{aaai25}

\begin{thebibliography}{10}
\providecommand{\url}[1]{#1}
\csname url@samestyle\endcsname
\providecommand{\newblock}{\relax}
\providecommand{\bibinfo}[2]{#2}
\providecommand{\BIBentrySTDinterwordspacing}{\spaceskip=0pt\relax}
\providecommand{\BIBentryALTinterwordstretchfactor}{4}
\providecommand{\BIBentryALTinterwordspacing}{\spaceskip=\fontdimen2\font plus
\BIBentryALTinterwordstretchfactor\fontdimen3\font minus \fontdimen4\font\relax}
\providecommand{\BIBforeignlanguage}[2]{{%
\expandafter\ifx\csname l@#1\endcsname\relax
\typeout{** WARNING: IEEEtran.bst: No hyphenation pattern has been}%
\typeout{** loaded for the language `#1'. Using the pattern for}%
\typeout{** the default language instead.}%
\else
\language=\csname l@#1\endcsname
\fi
#2}}
\providecommand{\BIBdecl}{\relax}
\BIBdecl

\bibitem{pgd}
A.~Madry, A.~Makelov, L.~Schmidt, D.~Tsipras, and A.~Vladu, ``Towards deep learning models resistant to adversarial attacks,'' \emph{arXiv preprint arXiv:1706.06083}, 2017.

\bibitem{fgsm_rs}
E.~Wong, L.~Rice, and J.~Z. Kolter, ``Fast is better than free: Revisiting adversarial training,'' \emph{arXiv preprint arXiv:2001.03994}, 2020.

\bibitem{n_fgsm}
P.~de~Jorge~Aranda, A.~Bibi, R.~Volpi, A.~Sanyal, P.~Torr, G.~Rogez, and P.~Dokania, ``Make some noise: Reliable and efficient single-step adversarial training,'' \emph{Advances in Neural Information Processing Systems}, vol.~35, pp. 12\,881--12\,893, 2022.

\bibitem{zero_grad}
Z.~Golgooni, M.~Saberi, M.~Eskandar, and M.~H. Rohban, ``Zerograd: Costless conscious remedies for catastrophic overfitting in the fgsm adversarial training,'' \emph{Intelligent Systems with Applications}, vol.~19, p. 200258, 2023.

\bibitem{fgsm_pgi}
X.~Jia, Y.~Zhang, X.~Wei, B.~Wu, K.~Ma, J.~Wang, and X.~Cao, ``Prior-guided adversarial initialization for fast adversarial training,'' in \emph{European Conference on Computer Vision}.\hskip 1em plus 0.5em minus 0.4em\relax Springer, 2022, pp. 567--584.

\bibitem{fgsm_uap}
C.~Pan, Q.~Li, and X.~Yao, ``Adversarial initialization with universal adversarial perturbation: A new approach to fast adversarial training,'' in \emph{Proceedings of the AAAI Conference on Artificial Intelligence}, vol.~38, no.~19, 2024, pp. 21\,501--21\,509.

\bibitem{ssat}
H.~Kim, W.~Lee, and J.~Lee, ``Understanding catastrophic overfitting in single-step adversarial training,'' in \emph{Proceedings of the AAAI Conference on Artificial Intelligence}, vol.~35, no.~9, 2021, pp. 8119--8127.

\bibitem{grad_align}
M.~Andriushchenko and N.~Flammarion, ``Understanding and improving fast adversarial training,'' \emph{Advances in Neural Information Processing Systems}, vol.~33, pp. 16\,048--16\,059, 2020.

\bibitem{gat}
G.~Sriramanan, S.~Addepalli, A.~Baburaj \emph{et~al.}, ``Guided adversarial attack for evaluating and enhancing adversarial defenses,'' \emph{Advances in Neural Information Processing Systems}, vol.~33, pp. 20\,297--20\,308, 2020.

\bibitem{nu_at}
------, ``Towards efficient and effective adversarial training,'' \emph{Advances in Neural Information Processing Systems}, vol.~34, pp. 11\,821--11\,833, 2021.

\bibitem{aaer}
R.~Lin, C.~Yu, and T.~Liu, ``Eliminating catastrophic overfitting via abnormal adversarial examples regularization,'' \emph{Advances in Neural Information Processing Systems}, vol.~36, pp. 67\,866--67\,885, 2023.

\bibitem{fgsm_mep_cs_fgsm_rs_cs}
M.~Zhao, L.~Zhang, Y.~Kong, and B.~Yin, ``Fast adversarial training with smooth convergence,'' in \emph{Proceedings of the IEEE/CVF International Conference on Computer Vision}, 2023, pp. 4720--4729.

\bibitem{fgsm_pco}
Z.~Wang, H.~Wang, C.~Tian, and Y.~Jin, ``Preventing catastrophic overfitting in fast adversarial training: A bi-level optimization perspective,'' in \emph{European Conference on Computer Vision}.\hskip 1em plus 0.5em minus 0.4em\relax Springer, 2024, pp. 144--160.

\bibitem{elle}
E.~A. Rocamora, F.~Liu, G.~G. Chrysos, P.~M. Olmos, and V.~Cevher, ``Efficient local linearity regularization to overcome catastrophic overfitting,'' \emph{arXiv preprint arXiv:2401.11618}, 2024.

\bibitem{liet}
C.~Pan, K.~Tang, Q.~Li, and X.~Yao, ``Mitigating catastrophic overfitting in fast adversarial training via label information elimination,'' in \emph{Proceedings of the IEEE/CVF International Conference on Computer Vision}, 2025, pp. 2991--3000.

\bibitem{robustbench}
F.~Croce, M.~Andriushchenko, V.~Sehwag, E.~Debenedetti, N.~Flammarion, M.~Chiang, P.~Mittal, and P.~Frossard, ``{RobustBench}: a standardized adversarial robustness benchmark,'' in \emph{Neurips 2020 Competition and Demonstration Track}.\hskip 1em plus 0.5em minus 0.4em\relax PMLR, 2021, pp. 141--151.

\bibitem{resnet}
K.~He, X.~Zhang, S.~Ren, and J.~Sun, ``Deep residual learning for image recognition,'' in \emph{Proceedings of the IEEE Conference on Computer Vision and Pattern Recognition}, 2016, pp. 770--778.

\bibitem{real_vit}
A.~Dosovitskiy, ``An image is worth 16x16 words: Transformers for image recognition at scale,'' \emph{arXiv preprint arXiv:2010.11929}, 2020.

\bibitem{wide_resnet}
S.~Zagoruyko and N.~Komodakis, ``Wide residual networks,'' \emph{arXiv preprint arXiv:1605.07146}, 2016.

\bibitem{new_1_bartoldson2024adversarial}
B.~R. Bartoldson, J.~Diffenderfer, K.~Parasyris, and B.~Kailkhura, ``Adversarial robustness limits via scaling-law and human-alignment studies,'' \emph{arXiv preprint arXiv:2404.09349}, 2024.

\bibitem{new_4_peng2023robust}
S.~Peng, W.~Xu, C.~Cornelius, M.~Hull, K.~Li, R.~Duggal, M.~Phute, J.~Martin, and D.~H. Chau, ``Robust principles: Architectural design principles for adversarially robust cnns,'' \emph{arXiv preprint arXiv:2308.16258}, 2023.

\bibitem{Rethinking_RobustBench}
C.~Pan, K.~Tang, Q.~Li, and X.~Yao, ``Rethinking robustbench: Is high synthetic-test data similarity an implicit information advantage inflating robustness scores?'' in \emph{2025 IEEE 12th International Conference on Data Science and Advanced Analytics (DSAA)}.\hskip 1em plus 0.5em minus 0.4em\relax IEEE, 2025, pp. 1--10.

\bibitem{auto_attack}
F.~Croce and M.~Hein, ``Reliable evaluation of adversarial robustness with an ensemble of diverse parameter-free attacks,'' in \emph{International Conference on Machine Learning}.\hskip 1em plus 0.5em minus 0.4em\relax PMLR, 2020, pp. 2206--2216.

\bibitem{cr_attack}
C.~Pan, Y.~Wu, K.~Tang, Q.~Li, and X.~Yao, ``Efficient robustness evaluation via constraint relaxation,'' in \emph{Proceedings of the AAAI Conference on Artificial Intelligence}, vol.~39, no.~6, 2025, pp. 6263--6271.

\bibitem{fgsm}
I.~J. Goodfellow, J.~Shlens, and C.~Szegedy, ``Explaining and harnessing adversarial examples,'' \emph{arXiv preprint arXiv:1412.6572}, 2014.

\bibitem{free_at}
A.~Shafahi, M.~Najibi, A.~Ghiasi, Z.~Xu, J.~Dickerson, C.~Studer, L.~S. Davis, G.~Taylor, and T.~Goldstein, ``Adversarial training for free!'' in \emph{Advances in Neural Information Processing Systems}, vol.~32, 2019, pp. 3353--3364.

\bibitem{wa}
P.~Izmailov, D.~Podoprikhin, T.~Garipov, D.~Vetrov, and A.~G. Wilson, ``Averaging weights leads to wider optima and better generalization,'' \emph{arXiv preprint arXiv:1803.05407}, 2018.

\bibitem{cifar10}
A.~Krizhevsky, G.~Hinton \emph{et~al.}, ``Learning multiple layers of features from tiny images,'' 2009.

\bibitem{deng2009imagenet}
J.~Deng, W.~Dong, R.~Socher, L.-J. Li, K.~Li, and L.~Fei-Fei, ``Imagenet: A large-scale hierarchical image database,'' in \emph{2009 IEEE Conference on Computer Vision and Pattern Recognition}.\hskip 1em plus 0.5em minus 0.4em\relax IEEE, 2009, pp. 248--255.

\bibitem{preresnet}
K.~He, X.~Zhang, S.~Ren, and J.~Sun, ``Identity mappings in deep residual networks,'' in \emph{European Conference on Computer Vision}.\hskip 1em plus 0.5em minus 0.4em\relax Springer, 2016, pp. 630--645.

\bibitem{label_smoothing}
M.~Goibert and E.~Dohmatob, ``Adversarial robustness via label-smoothing,'' \emph{arXiv preprint arXiv:1906.11567}, 2019.

\end{thebibliography}

\end{document}